\documentclass{article}

\usepackage{natbib}
\setcitestyle{square,sort,comma,numbers}
\usepackage{arxiv}
\usepackage{scalefnt}
\usepackage[utf8]{inputenc} 
\usepackage[T1]{fontenc}    
\usepackage{hyperref}       
\usepackage{url}            
\usepackage{booktabs}       
\usepackage{amsfonts}       
\usepackage{nicefrac}       
\usepackage{microtype}      
\usepackage{lipsum}
\newtheorem{definition}{Definition}
\newtheorem{propriete}{Property}
\usepackage{stmaryrd}
\usepackage{amsmath}
\usepackage{tikz}

\usetikzlibrary{decorations.pathreplacing}
\usepackage{tikz}

\tikzset{cross/.style={path picture={
  \draw[red]
    (path picture bounding box.south east)--(path picture bounding box.north west)
    (path picture bounding box.south west)--(path picture bounding box.north east);
}}}

\title{Intersection over Union is not a submodular function}

\author{
  KERDONCUFF Tanguy \\
  Univ Lyon, UJM-Saint-Etienne, CNRS,\\
  Institut d Optique Graduate School,\\
  Laboratoire Hubert Curien UMR 5516, F-42023,\\
  SAINT-ETIENNE, France\\
  \texttt{tanguy.kerdoncuff@laposte.net} \\
  \And
  EMONET Rémi \\
  Univ Lyon, UJM-Saint-Etienne, CNRS,\\
  Institut d Optique Graduate School,\\
  Laboratoire Hubert Curien UMR 5516, F-42023,\\ SAINT-ETIENNE, France\\
  \texttt{remi.emonet@univ-st-etienne.fr} \\
}

\begin{document}
\maketitle

\begin{abstract}
This short article aims at demonstrate that the Intersection over Union (or Jaccard index) is not a submodular function. This mistake has been made in \cite{yu2015lov} (Proposition 11) which is cited and used as a foundation in \cite{matthew2018lovasz}. The Intersection of Union is widely used in machine learning as a cost function especially for imbalance data and semantic segmentation.
\end{abstract}

\keywords{Intersection over Union \and Jaccard index \and Submodular \and Lovàsz extension}
\section{Definition}

We will start with the definitions of a submodular function and the Intersection over Union.

\begin{definition}
    With $m \in \mathbb{N}$ let $f : 2^{\{1,...,m\}} \rightarrow \mathbb{R}$. $f$ is submodular if and only if for all $\left( A,B \right) \in (2^{\{1,...,m\}})^{2}$ with $A \subseteq B$ and for all $x \in \llbracket 1,m\rrbracket \setminus A$.
    \begin{center}
        $f(A \cup \{x\}) - f(A) \geq f(B \cup \{x\}) - f(B)$
    \end{center}
    \label{def_submodular}
\end{definition}

\begin{definition}
    Let $(A,Y) \in (2^{\{1,...,m\}})^{2}$ and let $|A|$ be the function that return the number of element in $A$. While the definition is symmetric, we underline the fact that Y is fix in this problem.
    \begin{center}
        $IoU_{Y}(A) = \frac{|A \cap Y|}{|A \cup Y|}$
    \end{center}
    \label{def_IoU_ensemble}
\end{definition}

\section{The context around those research}

To minimize directly the Intersection over Union we have to find a differentiable loss function that have the same value as the IoU were the IoU is define. There is an infinite number of possible function so we are looking for special characteristics such as convexity. The Lovàsz approximation of a submodular function is convex. The Lovàsz approximation is also almost surely differentiable so it can be used as a loss function. Unfortunately, as demonstrated in Section \ref{Proof} the IoU is not submodular so its Lovàsz extension is not convex.

\section{Proof of the non-submodularity of the IoU}
\label{Proof}

We are going to demonstrate that the IoU is not submodular. Let $(A,B,Y) \in (2^{\{1,...,m\}})^{3}$ with $A \subset B$ and $|A \cap Y| = |B \cap Y| > 0$. Let $ab_{n} = |A \cap Y| = |B \cap Y|$, $a_{d} = |A \cup Y|$, $b_{d} = |B \cup Y|$, we also notice that $a_{d} < b_{d}$.
We want to demonstrate that for all $x \in 2^{\{1,...,m\}} \setminus (Y \cup B) $ we have $ R = (IoU_{Y}(A \cap \{x\}) - IoU_{Y}(A)) - (IoU_{Y}(B \cap \{x\}) - IoU_{Y}(B)) < 0$. Let $ x \in 2^{\{1,...,m\}} \setminus (Y \cup B) $

\begin{figure}
    \centering
    {\scalefont{5}
    \begin{tikzpicture}
    
    \draw[very thick, color=red] (0cm,0cm) rectangle (8cm,8cm);
    \draw[color=green] (5.2,4) circle (2.25cm);
    \draw[color={rgb:red,4;green,2;yellow,1}] (3cm,4) ellipse (2.5cm and 1cm);
    \draw[color=blue, dashed] (3.2cm,5cm) arc (90:180:1) ;
    \draw[color=blue, dashed] (3.2cm,3cm) arc (90:0:-1) ;

    \draw[rotate=-90, color=blue, dashed] (-3cm,3cm) arc[start angle=0,end angle=180, x radius=1, y radius=2.5];
    
    \draw[color=red] (2cm,7.3cm) node{$2^{\{1,...,m\}}$};
    \draw[color=green] (6cm,5cm) node{$Y$};
    \draw[color={rgb:red,4;green,2;yellow,1}] (1.3cm,4cm) node{$B$};
    \draw[color=blue] (2.6cm,4cm) node{$A$};
    \draw[color=purple] (4cm,4cm) node{$ab_{n}$};
    \draw (1.25cm,1.07cm) node{$x$};
    \draw[x=1.4ex,y=1.4ex,line width=.05ex] (1.5,0.8) -- (1.8,1.1) (1.5,1.1) -- (1.8,0.8);
    \end{tikzpicture}
    }
    \caption{Graphical representation of the sets in the particular case $ab_{n} = |A \cap Y| = |B \cap Y|$}
\end{figure}
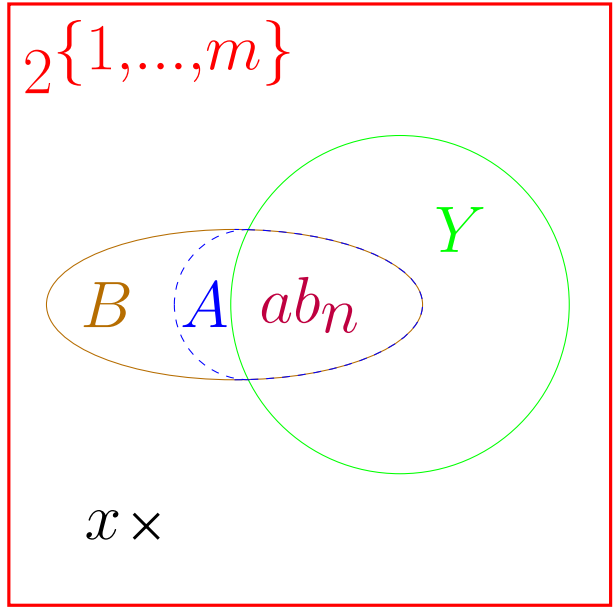

\begin{align}
    R &= \left( IoU_{Y}(A \cup \{x\}) - IoU_{Y}(A) \right) - (IoU_{Y}(B \cup \{x\}) - IoU_{Y}(B))\\
    & = \frac{|(A \cup \{x\}) \cap Y|}{|(A \cup \{x\}) \cup Y|} - \frac{|A \cap Y|}{|A \cup Y|} - \left( \frac{|(B \cup \{x\}) \cap Y|}{|(B \cup \{x\}) \cup Y|} - \frac{|B \cap Y|}{|B \cup Y|} \right) \\
    & = \frac{|A \cap Y|}{|A \cup \{x\} \cup Y|} - \frac{|A \cap Y|}{|A \cup Y|} - \left( \frac{|B \cap Y|}{|B \cup \{x\} \cup Y|} - \frac{|B \cap Y|}{|B \cup Y|} \right) \\
    & = \frac{ab_{n}}{a_{d} + 1} - \frac{ab_{n}}{a_{d}} - \left( \frac{ab_{n}}{b_{d} + 1} - \frac{ab_{n}}{b_{d}} \right) \\
    & = ab_{n} \left( \frac{a_{d} - a_{d} - 1}{(a_{d} + 1)a_{d}} - \frac{b_{d} - b_{d} - 1}{(b_{d} + 1)b_{d}} \right) \\
    & = ab_{n} \left( \frac{1}{(b_{d} + 1)b_{d}} - \frac{1}{(a_{d} + 1)a_{d}} \right) \\
\end{align}

And we have successively : 
\begin{align}
    0 < a_{d} & < b_{d} \\
    0 < (a_{d} + 1)a_{d} & < (b_{d} + 1)b_{d}\\
    \frac{1}{(b_{d} + 1)b_{d}} & < \frac{1}{(a_{d} + 1)a_{d}}\\
    \frac{1}{(b_{d} + 1)b_{d}} - \frac{1}{(a_{d} + 1)a_{d}} & < 0 \\
    ab_{n} \left( \frac{1}{(b_{d} + 1)b_{d}} - \frac{1}{(a_{d} + 1)a_{d}} \right) & < 0
\end{align}

Finally $R < 0$ with all the hypothesis on $x, A, B$, therefore \emph{the IoU is not submodular}.

In a similar way we take $x \in Y \setminus B$ to proves that -IoU is not submodular either.

\begin{align}
    R &= \left( IoU_{Y}(A \cup \{x\}) - IoU_{Y}(A) \right) - (IoU_{Y}(B \cup \{x\}) - IoU_{Y}(B))\\
    & = \frac{|(A \cup \{x\}) \cap Y|}{|(A \cup \{x\}) \cup Y|} - \frac{|A \cap Y|}{|A \cup Y|} - \left( \frac{|(B \cup \{x\}) \cap Y|}{|(B \cup \{x\}) \cup Y|} - \frac{|B \cap Y|}{|B \cup Y|} \right) \\
    & = \frac{|(A \cup \{x\}) \cap Y|}{|A \cup Y|} - \frac{|A \cap Y|}{|A \cup Y|} - \left( \frac{|(B \cup \{x\}) \cap Y|}{|B \cup Y|} - \frac{|B \cap Y|}{|B \cup Y|} \right) \\
    & = \frac{ab_{n} + 1}{a_{d}} - \frac{ab_{n}}{a_{d}} - \left( \frac{ab_{n} + 1}{b_{d}} - \frac{ab_{n}}{b_{d}} \right) \\
    & = \frac{1}{a_{d}} - \frac{1}{b_{d}} \\
\end{align}

And we have successively :

\begin{align}
    0 < a_{d} & < b_{d} \\
    \frac{1}{a_{d}} & > \frac{1}{b_{d}}\\
    \frac{1}{a_{d}} - \frac{1}{b_{d}} & > 0
\end{align}

Finally $R > 0$ if $x \in Y$, therefore \emph{-IoU is not submodular}.

\section{Explanation of the mistake}

We will highlight the error in \cite{yu2015lov}. The notation in \cite{yu2015lov} is different from the notation in this article because the demonstration is different. The equation (44) explains that :

\begin{propriete}
    Let $(A,B,Y) \in (2^{\{1,...,m\}})^{3}$ with $B \subset A$ and let $n_{A} = |Y \setminus A|$ and $n_{B} = |Y \setminus B|$.
    \begin{center}
        $n_{B} < n_{A}$
    \end{center}
    \label{prop_error}
\end{propriete}

This property is false because if $B \subset A$ then $|B| < |A|$ and $|Y \setminus A| < |Y \setminus B|$. The demonstration uses this false assertion to demonstrate that -IoU is submodular.

\bibliographystyle{unsrt}
\bibliography{IoU_is_not_submodular}

\end{document}